%% file: main.tex
\documentclass{IEEEtran}
\usepackage{cite}
\usepackage{amsmath,amssymb,amsfonts}
\usepackage{algorithmic}
\usepackage{graphicx}
\usepackage{caption}
\usepackage{booktabs}
\usepackage{multirow}
\usepackage{textcomp}
\usepackage[linesnumbered, ruled]{algorithm2e}
\usepackage{hyperref}
\usepackage{float}
\usepackage[pagewise]{lineno}
\SetKwRepeat{Do}{do}{while}%

\def\BibTeX{{\rm B\kern-.05em{\sc i\kern-.025em b}\kern-.08em
    T\kern-.1667em\lower.7ex\hbox{E}\kern-.125emX}}
\begin{document}
\title{Cross-modality Guidance-aided Multi-modal Learning with Dual Attention for MRI Brain Tumor Grading}
\author{Dunyuan Xu, Xi Wang*, Jinyue Cai and Pheng-Ann Heng, \IEEEmembership{Senior Member, IEEE}

\thanks{D. Xu, X. Wang, J. Cai and P. Heng are with the Department of Computer Science and Engineering, The Chinese University of Hong Kong, Hong Kong (email: dunyuanxu@cuhk.edu.hk, xiwang@cse.cuhk.edu.hk, jycai@cuhk.edu.hk, pheng@cse.cuhk.edu.hk)}
\thanks{X. Wang is also with Zhejiang Lab, Hangzhou, China and Department of Radiation Oncology, Stanford University School of Medicine, Palo Alto, CA, USA.}
\thanks{Asterisk indicates corresponding author.}
}

\maketitle

\begin{abstract}
Brain tumor represents one of the most fatal cancers around the world, and is very common in children and the elderly. Accurate identification of the type and grade of tumor in the early stages plays an important role in choosing a precise treatment plan.  The Magnetic Resonance Imaging (MRI) protocols of different sequences provide clinicians with important contradictory information to identify tumor regions. However, manual assessment is time-consuming and error-prone due to big amount of data and the diversity of brain tumor types. Hence, there is an unmet need for MRI automated brain tumor diagnosis. We observe that the predictive capability of uni-modality models is limited and their performance varies widely across modalities, and the commonly used modality fusion methods would introduce potential noise, which results in significant performance degradation. To overcome these challenges, we propose a novel cross-modality guidance-aided multi-modal learning with dual attention for addressing the task of MRI brain tumor grading. To balance the tradeoff between model efficiency and efficacy, we employ ResNet Mix Convolution as the backbone network for feature extraction. Besides, dual attention is applied to capture the semantic interdependencies in spatial and slice dimensions respectively. To facilitate information interaction among modalities, we design a cross-modality guidance-aided module where the primary modality guides the other secondary modalities during the process of training, which can effectively leverage the complementary information of different MRI modalities and meanwhile alleviate the impact of the possible noise. Experimental results on the BraTS2018 and BraTS2019 datasets demonstrate the effectiveness of the proposed method, outperforming the single modality-based approaches and several state-of-the-art multi-modal methods by a large margin, with an AUC of 0.985$\pm$0.019 and 0.966$\pm$0.021 on the two datasets, respectively.
\end{abstract}


\input{sections/introduction.tex}
\input{sections/relatedwork.tex}

\input{sections/method.tex}

\input{sections/experiments.tex}
\input{sections/discussion.tex}
\input{sections/conclusion.tex}

\bibliographystyle{IEEEtran}
\bibliography{reference}

\end{document}

%% file: sections/introduction.tex
\section{Introduction}
\begin{figure}[t]
    \centering
    \includegraphics[scale=0.23]{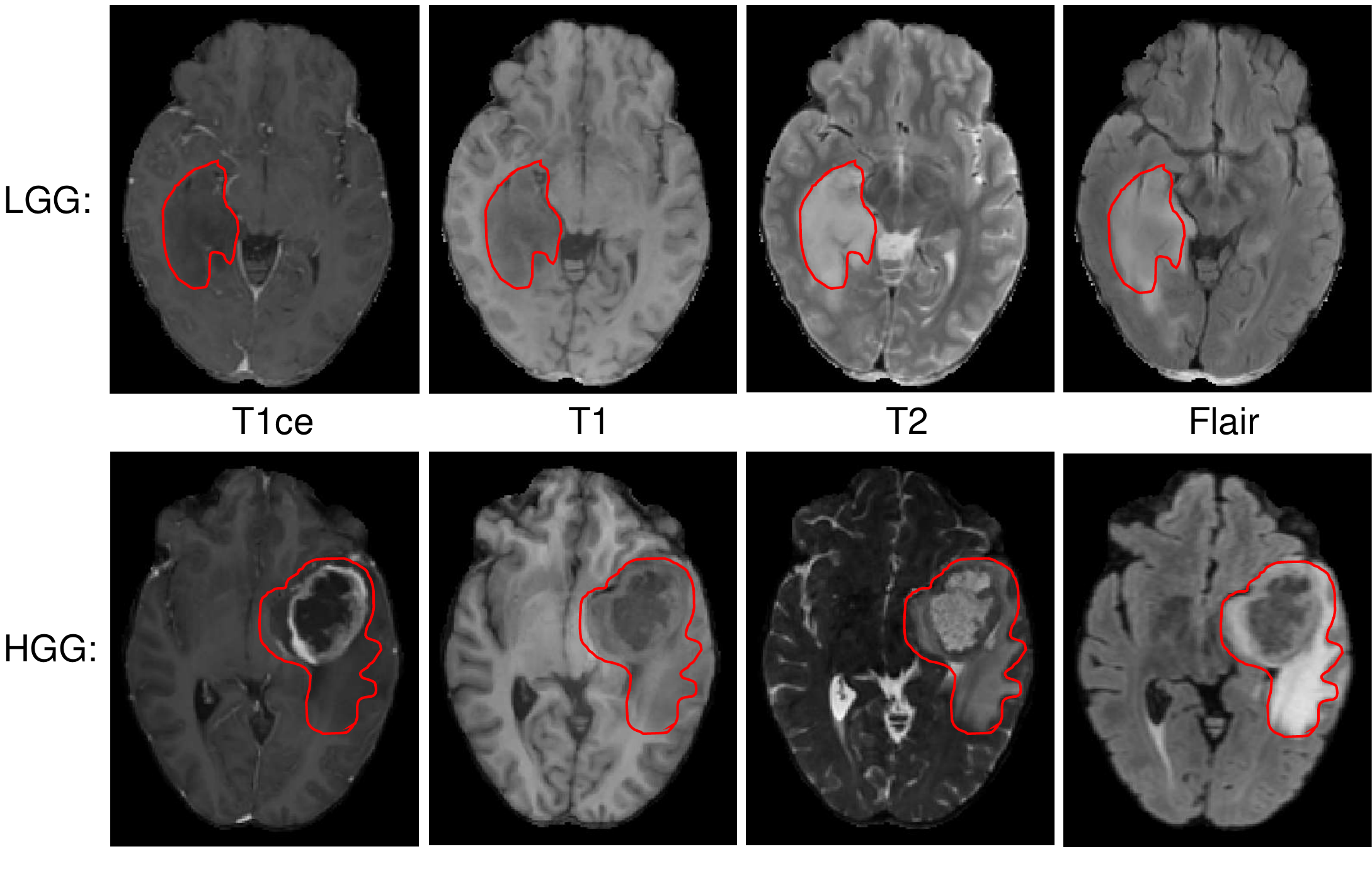}
    \caption{Examples of low-grade and high-grade cases in the T1ce, T1, T2, and Flair MRI images. These two cases are selected from BraTS2019. The whole tumor is enclosed in the red polygons.
    }
    \label{fig:four_modal}
\end{figure}
The brain tumor is one of the deadliest cancers due to its aggressive nature, heterogeneous characteristics, and low relative survival rate (According to statistics from the American Cancer Society, the survival rate of elderly patients with glioblastoma is only 6\%, and the expected overall survival period is 14.4 months \cite{r-5}). Among the primary brain tumors, gliomas are considered the most aggressive. As suggested by \cite{krafft2004analysis}, the World Health Organization divides brain tumors into the following 4 different grades according to the increasing degree of malignancy: pilocytic astrocytoma (grade I), fibrillary astrocytoma (grade II), anaplastic astrocytoma (grade III), and glioblastoma multiforme (grade IV). Scientists further classify these 4 types of brain tumor into two categories, Low Grade Glioma (LGG), which contains grade I and II brain tumors, and High Grade Glioma (HGG), which contains grade III and IV brain tumor \cite{lefkovits2022hgg}.
In general, high-grade gliomas proliferate and infiltrate the surrounding tissues at a great speed, and could carry a worse prognosis even if under proper treatment. Low-grade gliomas are less aggressive and tend to exhibit benign tendencies, but have a uniform recurrence rate and could progress to a high grade over time \cite{r-6}.

Early and accurate assessment of the brain tumor grade is crucial not only for estimating the prognoses of the patients but also for treatment planning and tumor growth evaluation. For HGG patients, adjuvant radiation therapy and chemotherapy after surgery have become the accepted routine practice that could improve the survival rate \cite{r-4}. On the contrary, as for LGG patient who is low-risk, radiation and chemotherapy are not necessary after surgery \cite{r-2} \cite{r-3}. Therefore, it is important to classify the tumor grade in advance.

Histopathological diagnosis of biopsy specimens is the gold standard for glioma grading \cite{r-1}. However, it is time-consuming, invasive, and prone to sampling error. The sampling error is often caused by the site of resection can not represent the highest malignancy in the biopsy sample as gliomas are heterogeneous \cite{r-8} (as only a small part of the entire tumor could be collected each time). In addition, there are several risk factors in the biopsy test, including cerebral hemorrhage and asymptomatic hemorrhage due to the biopsy needle \cite{r-7}, resulting in severe migraine, stroke, coma, and even death. Other associated risks include infection or seizures. Biopsy testing is shown to be slow and sometimes even dangerous, so it is necessary to find a timely, safe, and accurate method.

Magnetic Resonance Imaging (MRI), on the other hand, has the advantage of scanning the entire brain tumor in vivo and can display a strong association with the histopathological grade, which is a commonly used tool for the detection, diagnosis, and management of brain cancers in a non-invasive manner. Different modalities of MRI can be obtained by configuring the MRI scanner, such as T1-weighted (T1), T1-Post contrast-enhanced (T1ce), T2-weighted (T2), and T2-weighted fluid-attenuated inversion recovery (Flair), which have distinct functionalities. 
For instance, T1 facilitates the segmentation of tumors from healthy brain tissue. T1ce gives higher visibility of the tumor border. T2 shows the edema (fluid) around the brain tumor. Flair could be used for identifying edematous areas from the cerebrospinal fluid \cite{r-10}. Some typical MRI modality images of LGG and HGG is shown in Fig.~\ref{fig:four_modal}. In clinical routine practice, radiologists always combine multi-modal MRI images for decision-making~\cite{r-11}. However, manual assessment suffers from some shortcomings and challenges: (1) examining multiple MRI images is quite time-consuming; (2) it is also error-prone since some HGG may have some attributes of LGG, and vice-versa, which is a major challenge in accurate diagnosis of brain tumors even for experienced clinicians. Hence, there is an unmet need for automated brain tumor grading based on multi-modal MRI images.

Experts from the computer vision field have tried solving the brain tumor grading problem using Artificial Intelligence (AI) techniques. Some of them extracted radiomics features from tumor regions, which were then used to train traditional Machine Learning (ML) algorithms~\cite{cho2017classification}~\cite{cheng2020prediction}. With the success of deep learning in the field of medical image analysis, researchers employed different scales of deep Convolutional Neural Networks (CNNs) to extract more representative features from the tumor region for classification~\cite{liang2018multimodal}~\cite{ouerghi2022glioma}. Rather than tumor-level analysis, researchers also devoted great effort to fully-size image analysis in which either 2D or 3D CNN models were established~\cite{zhuge2020automated}~\cite{shahzadi2018cnn}. To take advantage of the complementary information of multiple modalities, different multi-modal approaches were also proposed~\cite{mai2022multimodal}~\cite{maneesha2019multi}.

Despite prominent progress that has been achieved by existing works, there are still some shortcomings and challenges: (1) the tumor-level analysis, either using traditional or deep learning features, relied on manually delineated tumor regions prior to classification, which largely prevented them from being fully automated; (2) handcrafted features, such as radiomics, always have limited representation, generally resulting in unsatisfactory results; (3) single modality-based methods could not always achieve influential performance given partial information encoded in one MRI image; (4) multi-modal analysis is favorable but also much more difficult due to inherent characteristics of MRI data (i.e., multiple modalities and high dimensionality). Therefore, model efficiency poses a big issue, which should be carefully considered, especially for real-world applications.

This paper proposes a novel guidance-aided multi-modal feature extraction algorithm that uses the extracted features from the \textit{primary} modality to guide the feature extraction process for the remaining \textit{secondary} modalities in which feature concatenation occurs between each of the secondary modalities and the primary modality. Experiment results show that this algorithm can extract more useful complementary information from the secondary modalities for assistance and enhancement while retaining the primary features to the greatest extent possible. On the other hand, due to the uneven distribution of a brain tumor across slices, different slices are not equally important, so the dual attention mechanism is introduced for forcing the model to focus on not only some specific slices but also informative regions in these slices, which makes the model stronger and more robust.
\par
Our main contributions can be summarized as follows:
\begin{enumerate}
    \item Through preliminary experiments, we discover that the diagnostic information encoded by different modalities varies greatly, and the general modality fusion strategies are not robust, even potentially resulting in considerable performance degradation.
    \item We propose a novel multi-modal framework with a lightweight ResNet Mix Convolution architecture to address the MRI brain tumor grading problem, which could achieve a good tradeoff between model efficiency and efficacy.
    \item We design a cross-modality guidance-aided module that could effectively leverage multiple modalities for training, in which the main contribution of the primary modality could largely be maintained. Besides, dual attention is applied to selectively emphasize features with more information while suppressing features with less information. In this manner, the model can better benefit from multi-modal complementary information. 
    \item Extensive experiments on two public datasets demonstrate the effectiveness of the proposed method. The importance of the essential components in the proposed method is also verified by ablation studies.
\end{enumerate}    

%% file: sections/relatedwork.tex
\section{Related Work}\label{relatedwork}

Extensive studies have been conducted for medical-related MRI analysis, including data pre-processing, model dimension diversity, single-modal leaning, and multi-modal learning. MRI data is the most common data source for diagnosing in a non-invasive way. Therefore, the analysis based on MRI images in different modalities attracts the growing attention of society. In this section, we focus on surveying this promising direction, especially the tumor-related MRI analysis as it is highly related to our work.

\subsubsection{MRI pre-processing}

MRI pre-processing conducts tumor-related classification tasks from two main aspects. One stream targets classifying the Region Of Interest (ROI) tumor region, which might be provided by the datasets.~\cite{ouerghi2022glioma}. Liang et al.~\cite{liang2018multimodal} step further to crop out the tumor region according to the segmentation ground truth before training. This could somehow be regarded as a kind of trick in the classification task. As in the practical situation, marking the region of the tumor area would be much more complicated than giving the label of tumor grade. To avoid using the tumor ROI ground truth during testing, some works try to train a U-net-like model to split the tumor beforehand ~\cite{pei2020brain}~\cite{zhuge2020automated}. 
However, these methods rely on the accuracy of the segmentation model, which is ineffective and imports more influencing noise factors.

Another significant stream of work is devoted to taking advantage of radiomics features to facilitate classification performance, such as shape, gray level, etc. Cheng et al.~\cite{cheng2020prediction}~\cite{cheng2021multimodal} apply 23 kinds of intensive methods to the original MRI image before extracting the radiomics features. However, during the extraction process, the mask ROI for the input MRI image is still required to improve the effectiveness of extracted features. The same situation appears in Ning et al.~\cite{ning2021multi} who use the provided tumor segmentation ground truth as the mask. That would be the same issue as the previous stream.


\subsubsection{Single-modality learning}

Previous work has already achieved some partial results on single-modal learning in medical image analysis, all approaches discussed in this subsection focusing on T1ce images. Seetha et al.~\cite{re-2} use a simple structured CNN network to achieve relatively satisfactory results. Similarly, Deepak et al.~\cite{re-3} employ transfer learning with a pretrained GoogleNet model. Díaz-Pernas~\cite{re-4} come up with a multi-scale CNN structure analyzed in three different spatial scales, which works just like the human visual system instead of only utilizing plane information. Other experts adopt Local Binary Patterns (LBP)~\cite{re-5} to analyze features for a single modality. Dutta et al.~\cite{dutta2022cdanet} propose dual attention could be utilized for investigating the T1ce MRI images in tumor segmentation tasks.
However, all methods described above only use the T1ce modality, leaving rich cross-modality data unexploited. Therefore, in this paper, we proposed a robust and accurate multi-modal model that could utilize complementary information for grading brain tumors from four different modalities (T1, T1ce, T2, and Flair).


\subsubsection{Multi-Modality Learning}
Multi-modality learning is the most commonly used method when coming across to dealing with tumor-related tasks. The highlighted parts of each modality are different, the accurate diagnosis conclusion should be made by comprehensive judgment through all modalities.  

Some experts use mathematics image processing methods, like Discrete Wavelet Transform (DWT) fusion~\cite{amin2020brain}, Non-Subsampled Contourlet Transform (NSCT) fusion~\cite{kaur2021multi}, Non-Subsampled Shearlet Transform (NSST) fusion~\cite{tan2020multimodal}, to pixel-fuse the MRI images from different modalities. Wang et al.~\cite{wang2022disentangled} propose using the group lasso penalty in the deep model to extract the MRI images’ complemented and redundant structure features, and deal with them separately. Almasri et al.~\cite{almasri2022artificial} use the CNN network combined with the hybrid optimization dynamic (HOD) algorithm to find the best fusion method. Hermessi et al.\cite{hermessi2021multimodal} suggest that pixel-level fusion is the basis of multi-modal learning, which provides support for higher-level methods.

To tackle the problem of unbalanced features in multi-modal learning, Liu et al.~\cite{liu2021contrastive} propose the Tuple Infomration  Noise-Contrastive Estimation (TupleInfoNCE) method which optimizes both the positive samples and negative samples and fuses their features to train a balanced model. Hu et al.~\cite{hu2021mmgcn} choose the bidirectional Long Short-Term Memory (LSTM) to fuse the different modalities and make predictions using the Graph Convolutional Network (GCN) network. Some experts found that the attention mechanism could be used in the feature fusion process. Wu et al.~\cite{wu2021multimodal} propose using Co-Attention after the feature extraction procedure could learn the dependencies between different features just like in the Natural Language Processing (NLP) field. Zhang et al.~\cite{zhang2022mmformer} deploy a transformer block after the linear projection of encoded features from different modalities to enhance the feasibility. The strength of feature-level fusion is that it suffers less information loss and it is less sensitive to noise compared with pixel-level fusion. 

Another important branch is decision-level fusion. The fusion process will be carried out to merge the predictions from every classifier. Chen et al. use a softmax layer to deal with the fused probability score from AlexNet and GoogleNet~\cite{chen2016xgboost}, then put the processed data into multiple different classifiers like Support Vector Machine (SVM), Linear Regression (LR), etc. Considering the different importance between modalities, Guo et al.~\cite{han2022multimodal} propose that each modality should have an independent weight during the fusion process. The decision level contains the least information from the original MRI image among all these three fusion metrics, it’s also the least sensitive to noise.

Multi-modality learning has already achieved satisfactory results, however, it also faces some challenges. Mai et al.~\cite{mai2022multimodal} suggest that the deep model might learn redundant knowledge from different modalities, which is out of sense. Some samples may even introduce more unwanted noise than useful information. This opinion is also supported by Maneesha et al.~\cite{maneesha2019multi}. Han et al.~\cite{han2022multimodal} further raise a question on how trustworthy are for these multi-modal fusion methods, especially for medical diagnosis. In addition, with the increasing number of different modalities, the network structure will become more and more complex, which could cause an over-fitting problem~\cite{wang2020makes}. In a word, using complementary information from different modalities and avoiding information redundant has become the key challenge in multi-modal learning in the medical image field.

In this paper, we adopt feature-level fusion and exploit the advantages of it. Our work overcomes the problems existing in the methods mentioned above and performs better since: 
a) Our model does not require the segmentation region for dealing with the input data. 
b) Our model does not need complicated data pre-processing steps like intensive transform. 
c) Our model shows more robustness and achieves higher scores under the three-fold cross-validation.

%% file: sections/method.tex
\section{Method}\label{method}
\begin{figure*}[th!]
    \centering
    \includegraphics[scale=0.6]{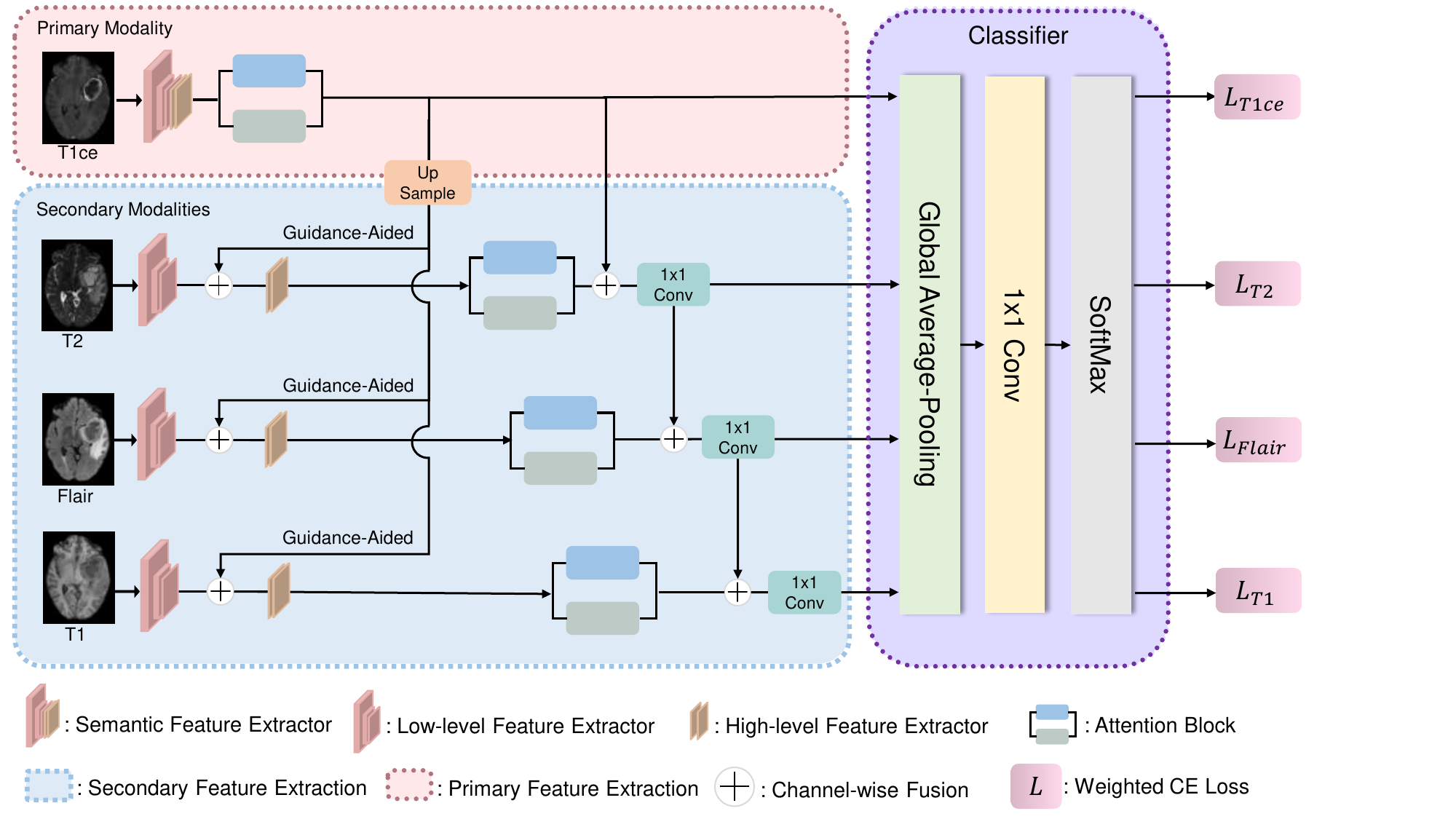}
    \caption{The primary high-level feature is extracted by a \textit{Semantic Feature Extractor (SFE)} followed by the \textit{dual attention} module and is upsampled to guide the secondary feature extraction process. In secondary modality feature extraction, the low-level features extracted by a \textit{Low Feature Extractor (LFE)} will be concatenated with the upsampled guidance features to achieve cross-modality guidance. The secondary high-level features are then extracted by a \textit{High Feature Extractor (HFE)} and merged with the cross-modality feature in the previous step. We calculate the cross-entropy loss on each prediction for every high-level feature combination. A 1×1 convolution after each feature concatenation is applied to make the number of features identical, and compatible with the setting of the classifier. 
    }
    \label{fig:main}
\end{figure*}
\label{sec:pagestyle}
In this work, we propose a novel multi-modal framework to address the task of MRI brain tumor grading. As shown in Fig~\ref{fig:main}, the proposed method adopts residual neural network (ResNet) mixed convolution proposed in~\cite{chatterjee2022classification} as the backbone network to extract representative features, which can largely reduce the computational cost. To improve the discrimination capability of the model, we design a cross-modality guidance-aided module to enable information interaction among modalities, which can effectively leverage the complementary information of different MRI modalities. Here, the primary modality is used to guide the secondary modalities during feature extraction. To enhance the discriminant ability of feature representations for classification, we embed the dual-attention module for aggregating long-range contextual information afterward. We will elaborate on each of the essential components in detail.


\begin{table*}[h]
\centering
\caption{The detailed network structure of RMC model.}
\small
\renewcommand\arraystretch{1.5}
\setlength{\tabcolsep}{0.8mm}{
\begin{tabular}{|c|c|c|c|c|c|c|}
\hline
3D Block       & 3D Block             & 2D Block                                                                       & 2D Block                                                                       & 2D Block                                                                       & Avgpool         & FC    \\ \hline
kernel: (3$\times$7$\times$7) & \{kernel: (3$\times$3$\times$3)\}$\times$4 & \begin{tabular}[c]{@{}c@{}}\{kernel: (1$\times$3$\times$3)\}$\times$4\\ kernel: (1$\times$1$\times$1)\end{tabular} & \begin{tabular}[c]{@{}c@{}}\{kernel: (1$\times$3$\times$3)\}$\times$4\\ kernel: (1$\times$1$\times$1)\end{tabular} & \begin{tabular}[c]{@{}c@{}}\{kernel: (1$\times$3$\times$3)\}$\times$4\\ kernel: (1$\times$1$\times$1)\end{tabular} & kernel: (1$\times$1$\times$1) & 512$\times$2 \\ \hline
\end{tabular}
}
\label{RMC model}
\end{table*}

\subsection{ResNet mixed convolution}
\label{ssec:subhead}
\begin{figure}[h]
    \centering
    \includegraphics[scale=0.26]{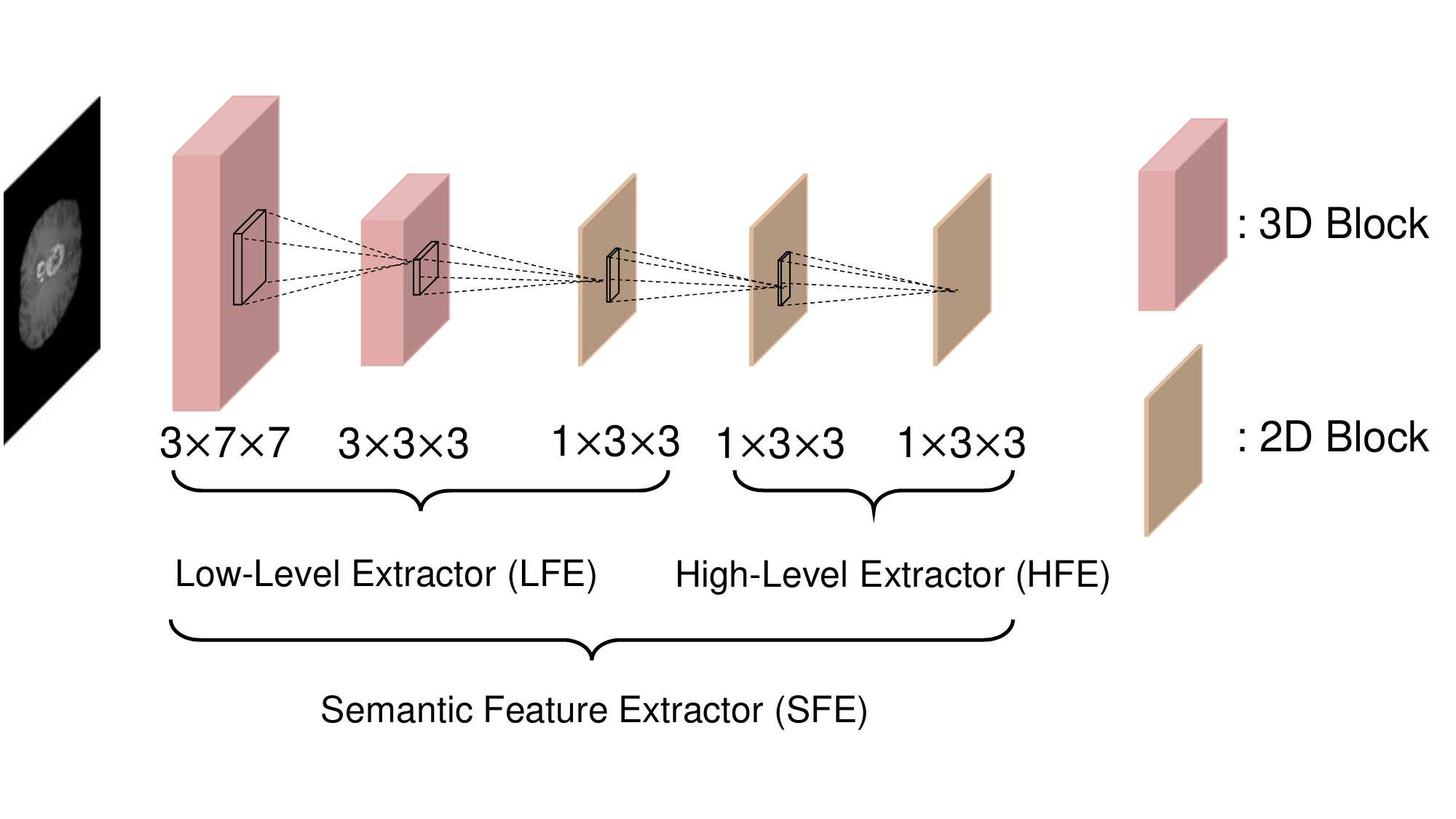}
    \caption{The feature extraction part of the RMC model, which makes up the Semantic Feature Extractor (SFE), further divided into a Low-level Feature Extractor (LFE) and a High-level Feature Extractor (HFE).}
    \label{fig:encoder}
\end{figure}

ResNet Mixed Convolution (\textbf{RMC}), by considering the slices as a sequence of images over time, was originally proposed in~\cite{chatterjee2022classification} to enable learning specific spatial and temporal relationships while reducing computational costs, which shows superior performance to the pure 3D ResNet and another spatiotemporal model ResNet (2+1)D on the task of brain tumor classification~\cite{chatterjee2022classification}. Here, we take RMC as the baseline method and adopt RMC as the backbone network in our framework for feature extraction. As shown in Fig.~\ref{fig:encoder}, the main body of ResNet Mixed Convolution consists of a combination of 2D and 3D Convolutions, including a stem containing a 3D convolution layer, then one 3D convolution block, and three 2D convolution blocks. We kept the configurations, such as kernel size, stride, and padding, identical to the original settings (See Table~\ref{RMC model}).

From the preliminary experiments on the baseline method, we observed that the performance of models trained with different modalities varies significantly under the same training conditions (For more details, please refer to Section~\ref{comparison}), and the simple model ensemble on averaging predicted probabilities leads to a worse result. The underlying reason might be ascribed to the fact that the tumor information enclosed in different modalities is distinct. Hence, a single modality is hardly sufficient for accurate diagnosis without other information being completed. We also experimentally verify that modality fusion without any regularization (e.g. pixel, feature, or decision fusion) could result in worse outcomes, implying that a simple fusion could potentially introduce noise that suppresses the information that is originally useful. To alleviate this problem, we enlisted two tactics: (i) dual attention, and (ii) cross-modality guidance.

\subsection{Dual attention}
\label{ssec:subhead}
Since the introduction of the attention mechanism \cite{vaswani2017attention}, this technique has been widely used in many kinds of AI-aided tasks in computer vision fields in recent years, and it has shown its strong ability on different kinds of tasks. \textit{Dual attention} \cite{fu2019dual} is another research outcome developed based on the self-attention mechanism, which calculates the attention on both the channel and the position dimension respectively. 

In the MRI brain tumor grading task, two influencing factors need to be considered: the tumor position, and the tumor size. The location of tumor growth is generally random, so the model is supposed to focus on learning from some specific position to distinguish the tumor types. Meanwhile, the brain tumor might have an irregular 3D structure, and the tumor size contained in each slice is different, which infers that some slices are more important for grading. In this regard, the general idea of dual attention can be applied in this study to calculate the attention map for both position and slice dimensions.

As shown in Fig.\ref{fig:main}, the dual attention is inserted into the feature extraction module to amplify or suppress different regions in the MRI volumes. The two types of attention are calculated simultaneously. For more details regarding dual-attention, please refer to Ref.~\cite{fu2019dual}.

\begin{figure}[h]
    \centering
    \includegraphics[scale=0.28]{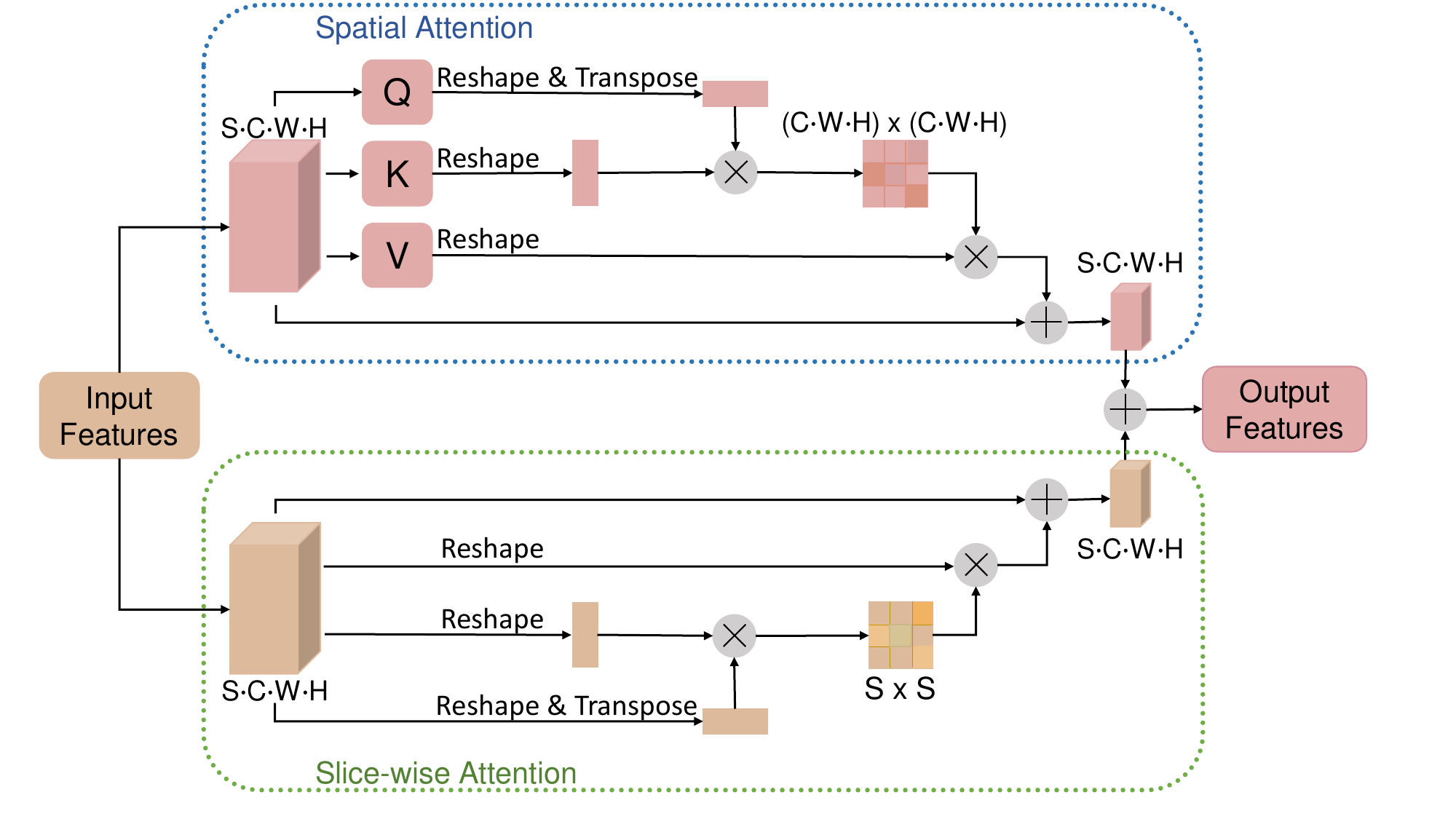}
    \caption{Dual attention structure. Calculate the respective attention for the spatial (channel$\times$height$\times$width) and slice (slice$\times$slice) then combine their results together.}
    \label{fig:attention}
\end{figure}

\subsection{Guidance-aided feature extraction module}
\label{ssec:subhead}
Ideally, the model trained using more modalities should perform better in brain tumor grading since more complementary information will be imported with the increasing of data. As mentioned in Section~~\ref{comparison}, there exists a large difference among the uni-modality models. According to the classification performance achieved by the four modalities, we define two concepts here: 
\begin{itemize}
\item \textbf{Primary modality}: the modality that performs the best in uni-modal training (e.g. in the BraTS dataset, the primary modality is T1ce.)
\item \textbf{Secondary modality}: the rest of the modalities in the whole dataset other than the primary modality (e.g. in the BraTS dataset, the secondary modalities are Flair, T1, and T2). Note that the secondary modalities are subsequently processed in descending order according to their performance on the baseline method.
\end{itemize}


For better mining extra complementary and beneficial information of the secondary modalities, while maintaining as much important information of the primary modality as possible, we propose a cross-modality guidance-aided multi-modal learning module. The guidance from the primary module is believed to encourage the model to selectively learn from the complementary regions while eliminating the potential noise caused by joint learning.


In general, features extracted by hidden layers of the CNN network present a hierarchical structure, from local to global specifications of the objects. Usually, the features from the shallow layers that are closer to the input are more general, referred to as \textit{low-level features}, which depict the detailed local patterns of the object, such as color, corner, edge et al. The features from the deep layers that are closer to the output are more task-specific, referred to as \textit{high-level features}, which describe the holistically global overview of the object. Compared to low-level features, high-level features encode more semantic information and thus are more representative and expressive.

As shown in Fig~\ref{fig:featuremap}, there exists a huge semantics gap between the high-level features of the primary modality and the raw input MRI images of the secondary modalities, the cross-modality guidance from the primary modality to the secondary modality might be too abstract and ambiguous. To this end, we propose to impose the guidance from the \textit{high-level} features of the primary modality and the \textit{low-level} features of the secondary modality rather than the raw input. Specifically, we employ the blocks before the pooling layer of the RMC model as the \textit{semantic feature extractor} (SFE), which is further divided into two parts: a \textit{low-level feature extractor} (LFE), and a \textit{high-level feature extractor} (HFE). More specifically, the first three blocks of the SFE constitute the LFE, which extracts the low-level features that contain more detailed information more specifically. The last two blocks comprise the high-level feature extractor, which extracts the high-level semantic features that are more representative and expressive. To selectively emphasize features that are more informative while suppressing
features with less information in the MRI volume, dual attention is further applied to the outcomes of these extractors.


The whole procedure of the proposed guidance-aided multi-modal feature extraction is composed of two parts: the primary modality feature extraction, and the secondary modality feature extraction. In the primary modality feature extraction module, where the SFE is used to generate feature maps of the primary modality image, followed by the dual attention operation, yielding the high-level feature representation $f_{p}^{h}$. 
A subsequent classifier is to make the prediction for the attention result. 
Besides, the high-level features of the primary modality are regarded as instructive information, which will be used as guidance in the feature extraction for the secondary modalities.

In the secondary modality feature extraction module, for $i$-th secondary modality, $i=1,2,3$, the low-level feature extractor is first employed to extract the low-level features $f_{s,i}^{l}$. The cross-modality guidance is achieved by concatenating the upsampled high-level features of the primary modality $\hat{f}_{p}^{h}$ with the low-level features of the $i$-th secondary modality $f_{s,i}^{l}$ by channel-wise fusion. 
Next, the concatenated features $\{f_{s,i}^{l},\hat{f}_{p}^{h}\}$ are passed through the high-level feature extractor to generate the features maps of the secondary modality. Likewise, dual attention is further applied to exploit the feature's importance. The ultimate high-level features of the $i$-th secondary modality are denoted by $f_{s,i}^{h}$. 

After the representative high-level features of all modalities are obtained, $f_{p}^{h}$ for the primary modality and $f_{s,i}^{h}$, $i=1,...,M$ for the secondary modalities, we propose a cumulative learning strategy for modality fusion. In particular, we assign each of the four modalities with different levels of priority according to their performance on the baseline methods (i.e., T1ce has the highest priority, followed by T2, Flair, and T1. We assume that the higher priority, the most information encoded in the modality. With the highest priority, the primary modality is processed first. Its high-level features are directly fed into a classification head $\mathcal{G}$ to yield the prediction. For simplicity, we remove the subscription $h$ from all denotations of the high-level feature (i.e., $f$) hereafter, unless otherwise specified.

Since the class distribution of the Brain Tumor Segmentation (BraTS) datasets is unbalanced, the weighted cross-entropy loss is applied here for controlling the prediction output. 
The weighted cross entropy loss is formulated as follows:

\begin{equation}
    \mathcal{L}_{\rm p} = -\sum\limits_{j}\omega_j[y_j*log(\mathcal{G}(f_{p}^{j}))],
\label{eq:wce}
\end{equation}

where $y_j$ means the true distribution of class $j$ and $\mathcal{G}(f_{p}^{j})$ represents the estimated distribution of class $j$, and $\omega_j$ denote the class weight, which is computed by:

\begin{equation}
    \omega_j = 1 - \frac{S_j}{\sum_{c\in C}^{} S_c} ,
\label{eq:omega}
\end{equation}
where the $S_j$ is the number of samples in class $j$, $C$ represents the total number of different classes.

As the secondary modalities would provide extra information, gradually integrating the additional information of the secondary modalities into the primary modality in the descending order of priority would greatly contribute to knowledge accumulation. Thus, when it comes to the $i$-th secondary modality, we fuse its high-level features with the features of modalities with a higher priority. It is worth noting that to reduce the computation cost, we apply a $1\times 1$ convolution after each feature concatenation to make the number of features identical, and compatible with the setting of the classification head $G$. Similarly, we apply the weighted cross-entropy loss over the $i$-th secondary modality:

\begin{equation}
    \mathcal{L}_{\rm s,i} = -\sum\limits_{j}\omega_j[y_j*log(\mathcal{G}(t_{i}^{j}))],
\label{eq:wce2}
\end{equation}

\begin{equation}
    t_{i} = \mathcal{T}(\{t_{i-1},f_{s,i}\}),
\label{eq:gi}
\end{equation}

\begin{equation}
    t_{1}=\mathcal{T}(\{f_{p},f_{s,1}\}),
\label{eq:g1}
\end{equation}
where $\{,\}$ denotes feature concatenation, and $\mathcal{T}$ denotes $1\times 1$ convolution that changes the number of feature maps. 

The overall objective function is formulated as:

\begin{equation}
\mathcal{L}_{\rm total} = \mathcal{L}_{\rm p}+ \sum_{i=1}^{M}\mathcal{L}_{\rm s,i},
\label{eq:total}
\end{equation}
where $M$ is the number of the secondary modalities, i.e., 3, in this work.

\begin{algorithm}[h]

\caption{Cross-Modality Guidance Algorithm}
\label{alg:algorithm}

\setlength\abovedisplayskip{10pt}
\setlength\belowdisplayskip{10pt}
\KwIn{ Numbers of training iterations $T_{p}$, $T_{s}$;
\par \quad Primary samples $P = \{X_{p}, y_{p}; p=(1,...,N_{p})\}$; 
\par \quad Secondary samples $S = \{X_{s,i}, y_{s,i}; s=(1,...,N_{s})\}$;
\par \quad Secondary modalities $I = [T1, T2, Flair]$.
 \par \quad SFE  $F_P$; LFE  $F_{s,i}^{l}$; HFE  $F_{s,i}^h$
}

\KwOut{ SFE parameters  $\theta_{p}$; Classifier parameters $\theta_{c}$;
\par \quad   LFE parameters  $\theta_{s,i}^{l}$ for i $\in$ I;

\par \quad   HFE parameters  $\theta_{s,i}^{h}$ for i $\in$ I.
}

\BlankLine

Initialize network  $F_P(\theta_{p})$, $F_C(\theta_{c})$
	
\For{p=1, $\cdots$, $T_{p}$}{

            Primary high-level feature ${f}_{p}^{h}$ $\gets$  $F_P(y_p|X_{p}; \theta_{p})$
            
            Calculate ${\mathcal{L}}_{\rm P}$ with ${f}_{p}^{h}$, $y$ using Eq.\ref{eq:wce} \ref{eq:omega}
            
            Update $\theta_{p}, \theta_{c}$ with  $\mathcal{L}_{\rm P}$

  }

\For{i in I}{
    
    $F_{s,i}^{l}(\theta_{s}^{l})$ $\Leftarrow$ CNN block 1, 2, 3 from $F_P(\theta_{p})$
        
    $F_{s,i}^{h}(\theta_{s}^{h})$ $\Leftarrow$ CNN block 4, 5 from $F_P(\theta_{p})$
}
    
\For{q=1, $\cdots$, $T_{s}$}{

    \For{i in I}{
    
        ${f}_{p}^{h}$ $\gets$  $F_P(y_p|X_{p}; \theta_{p})$
        
        $\hat{f}_{p}^{h}$ = $upsample({f}_{p}^{h})$
        
        Low-level feature $f_{s,i}^{l}$ $\gets$ $F_{s,i}^{l}(y_{s,i}|X_{s,i}; \theta_{s,i}^{l})$
        
        Guidance-aided $f_G$ $\gets$ $concat\{f_{s,i}^{l}, \hat{f}_{p}\}$;
    
        High-level feature $f_{s,i}^{h}$ $\gets$ $F_{m}^{h}(f_G; \theta_{m}^{h})$
        
         Calculate ${\mathcal{L}}_{\rm s,i}$ with $f_{s,i}$,  $f_{p}$, $y$ using Eq.\ref{eq:wce2} \ref{eq:gi} \ref{eq:g1}
    
        Calculate ${\mathcal{L}}_{\rm total}$ with ${\mathcal{L}}_{\rm P}$, ${\mathcal{L}}_{\rm s,i}$  using Eq.\ref{eq:total}
    
    }
                
    Update $\theta_{s,i}^{h}$, $\theta_{s,i}^{l}$with ${\mathcal{L}}_{\rm total}$

}

\end{algorithm}

%% file: sections/experiments.tex
\section{Experiments and Results}\label{experiments}
\subsection{Dataset}
We use two public datasets, BraTS2018 and BraTs2019~\cite{menze2014multimodal}~\cite{ bakas2017advancing}~\cite{bakas2018identifying} to evaluate the proposed method. As listed in Table~\ref{tab:datasets}, the BraTS2018 dataset includes 285 patients with brain gliomas (210 HGG and 75 LGG), and the BraTS2019 dataset includes 335 patients with 259 HGG and 76 LGG, respectively. Each patient has taken four modalities of 3D MRIs (T1, T1ce, T2, and Flair) that are rigidly aligned pixel by pixel. Two typical MRI cases (i.e., an LGG case and an HGG case) are selected from BraTS2019 and shown in Fig.~\ref{fig:four_modal}. All images are skull-stripped and stored in nii.gz format. We use the open-source python library SimpleITK~\cite{lowekamp2013design} for resolving the data. The spacing for each MRI is 1$\times$1$\times$1 $mm^3$ isotropic resolution and with the pixel resolution of 240$\times$240$\times$155. The classification ground-truth labels (i.e., LGG and HGG) are provided. Besides, the tumor regions tumor are delineated by experts, which include three tumor sub-regions: the enhancing tumor (ET), the tumor core (TC), and the whole tumor (WT). Although the proposed method is established upon the classification labels only, the provided pixel-wise annotations of the tumor region allow for comparison with existing approaches that rely on tumor-wise analysis.


\begin{table}[h]
\centering
\caption{Summary of BraTS 2018 and BraTS 2019 including the number and proportion of different tumor categories in each dataset.}
\small
\renewcommand\arraystretch{1.5}
\setlength{\tabcolsep}{7mm}{
\begin{tabular}{ccc}
\hline
      & BraTS2018     & BraTS2019     \\ \hline
HGG   & 210 (73.68\%) & 259 (77.31\%) \\
LGG   & 75 (26.32\%)  & 76 (22.69\%)  \\ \hline
Total & 285           & 335           \\ \hline
\end{tabular}
}
\label{tab:datasets}
\end{table}


\subsection{Evaluation metrics}
We assess the model performance with four metrics: i) Area under the ROC curve (\textbf{AUC}); ii) \textbf{Accuracy}; iii) \textbf{Sensitivity}, true positive rate (TPR), refers to the probability of a positive test, conditioned on truly being positive; iv) \textbf{Specificity}, true negative rate (TNR), refers to the probability of a negative test, conditioned on truly being negative. The threshold value for computing accuracy, sensitivity, and specificity is the default 0.5.


For a comprehensive evaluation, three-fold cross-validation is applied to all experiments on the two datasets. The whole dataset is randomly divided into three folds evenly. In each training round, two folds constitute the training set and the rest one is the test set. We calculate the mean and standard deviation for these three folds, and the scores are represented in the form of \textit{mean}$\pm$\textit{std}.

\subsection{Implementation details}
TITAN-RTX3090 GPU. We used the Adam optimizer\cite{kingma:adam} for updating parameters in the network with weight decay = 0.001. The MRI images were cropped to 150$\times$240$\times$240, while the original resolution is 155$\times$240$\times$240. The batch size was 1. We applied the data augmentation including horizontal flipping with a probability of 25\%, random affine with a degree in [-10, 10] scale in [0.9, 1.2], and rescale intensity with percentile in [0.5,99.5]. In this study, the weight coefficient $\omega$ in Eq.\ref{eq:omega} for the weighted cross entropy loss was initialized as  [2.8:1] and [3.4:1] for [LGG:HGG] on the BraTS2018 and the BraTS2019 respectively.

We employed our framework in two stages. The learning rate remains the same during the training process and was initialized to 0.00001, the dropout rate was 0.5 for both stages. The primary modality was T1ce, secondary modalities were T1, T2, and Flair. In the first stage, the semantic feature extractor (SFE) was trained with the T1ce MRI images for 200 epochs in order to obtain a high-performance SFE.  In the second stage, secondary modalities were trained under multi-modal, including T1ce, T1, T2, and Flair, the training epoch is 50. The training process takes around 12 hours for stage one and 10 hours for stage two.

\subsection{Experiments on the uni-modality models}
\label{comparison}
Given the four modalities of MRI images available, it is worth investigating the predictive power of each uni-modality model for brain tumor grading. Hence, we employed the baseline method, pre-trained ResNet Mix Convolution~\cite{chatterjee2022classification}, as the classifier, and took one modality as input each time, resulting in 4 uni-modality models for each dataset, referred to as T1-model, T1ce-model, T2-model, and Flair-model, respectively. As shown in Table~\ref{single18} and Table~\ref{single19}, we observe that results from different modalities vary greatly. The model trained with T1ce modality (i.e., T1ce-model) achieves the best AUC value on the two datasets (i.e., 0.917 and 0.901 on BraTS2018 and BraTS2019), which surpasses the other three models by a large margin. Although the T2-model outperforms slightly better than Flair-model does, the classification performance is far unsatisfactory. Among all uni-modality models, T1-model presented the worst results, with an AUC of 0.756 and 0.733 on the BraTS2018 and BraTS2019 datasets, respectively. Furthermore, We use the combined prediction of the four uni-modality models to build a final ensemble model, in which the probabilities yielded by all different uni-modal models are averaged. The corresponding results are listed in the last row in Table~\ref{single18} and Table~\ref{single19}. Surprisingly, the average performance of the four models is even worse than any uni-modality model. 

From the above observation, we can infer that different modalities carry distinct diagnostic information, and without any regularization, the model ensemble would even degrade the performance of the models. The possible reason behind is that model averaging could introduce substantial noise that suppresses the useful information of individual modalities. These preliminary experiments lay a solid foundation of this study, which motivates us to develop an effective multi-modal approach to fully exploit the complementary information from different modalities for improving the classification performance. 



\begin{table}[h]
\centering
\caption{The RMC model training with different uni-modal and ensemble results on BraTS2018.}
\small
\renewcommand\arraystretch{1.5}
\setlength{\tabcolsep}{1mm}{
\begin{tabular}{ccccc}
\hline
Modality & AUC   & Accuracy & Sensitivity & Specificity \\ \hline
T1ce      & \textbf{0.917$\pm$0.061}  & \textbf{0.926$\pm$0.038} & \textbf{0.933$\pm$0.046}  & \textbf{0.900$\pm$0.075}     \\
Flair    & 0.798$\pm$0.043 & 0.804$\pm$0.061  & 0.787$\pm$0.12     & 0.809$\pm$0.036     \\
T1       & 0.756$\pm$0.027 & 0.818$\pm$0.024  & 0.627$\pm$0.046   & 0.886$\pm$0.029   \\
T2       & 0.803$\pm$0.021 & 0.817$\pm$0.037  & 0.773$\pm$0.083   & 0.833$\pm$0.073           \\
Ensemble       & 0.642$\pm$0.062 & 0.805$\pm$0.038  & 0.293$\pm$0.113   & 0.990$\pm$0.011        \\ \hline
\end{tabular}
}
\label{single18}
\end{table}

\begin{table}[h]
\centering
\caption{The RMC model training with different uni-modal and ensemble results on BraTS2019.}
\small
\renewcommand\arraystretch{1.5}
\setlength{\tabcolsep}{1mm}{
\begin{tabular}{ccccc}
\hline
Modality & AUC   & Accuracy & Sensitivity & Specificity \\ \hline
T1ce     & \textbf{0.901$\pm$0.023}  & \textbf{0.911$\pm$0.031} & \textbf{0.882$\pm$0.037} & \textbf{0.919$\pm$0.041}     \\
Flair    & 0.770$\pm$0.043 & 0.773$\pm$0.032  & 0.765$\pm$0.136     & 0.776$\pm$0.069     \\
T1       & 0.733$\pm$0.038 & 0.839$\pm$0.015  & 0.539$\pm$0.086   & 0.927$\pm$0.017   \\
T2        & 0.805$\pm$0.019 & 0.836$\pm$0.042  & 0.749$\pm$0.064   & 0.861$\pm$0.069       \\ 
Ensemble       & 0.651$\pm$0.021 & 0.834$\pm$0.021  & 0.314$\pm$0.016   & 0.988$\pm$0.025     \\ \hline
\end{tabular}
}
\label{single19}
\end{table}

\subsection{Ablation study}
We conduct a series of ablation studies to investigate the effectiveness of the two essential components in the proposed method: dual attention and the guidance-aided feature extraction module. 

\subsubsection{Effectiveness of dual attention}
We inserted the dual attention module into the baseline network and carried out experiments using the T1ce modality MRI images on the BraTS2018 and BraTS2019 datasets. The corresponding results are reported in Table~\ref{DA_2018} and Table~\ref{DA_2019}, accordingly. It is noticeable that with the involvement of dual attention, the performance of the model is considerably boosted, improving the AUC by 2.4-2.9\% in absolute terms. The experiment results are in line with our expectations, suggesting that the dual attention module can indeed help the network capture more important features for grading brain tumors.

\begin{table}[h]
\centering
\caption{Comparison between RMC and DA\_RMC methods on the BraTS2018 dataset under T1ce modality}
\small
\renewcommand\arraystretch{1.5}
\setlength{\tabcolsep}{1mm}{
\begin{tabular}{ccccc}
\hline
Method  & AUC   & Accuracy & Sensitivity & Specificity \\ \hline
RMC      & 0.917$\pm$0.061  & 0.926$\pm$0.038 & \textbf{0.933$\pm$0.046}  & 0.900$\pm$0.075        \\
DA\_RMC & \textbf{0.941$\pm$0.020}  & \textbf{0.940$\pm$0.005} & 0.918$\pm$0.043  & \textbf{0.947$\pm$0.022}    \\ \hline
\end{tabular}
}
\label{DA_2018}
\end{table}


\begin{table}[h]
\centering
\caption{Comparison between RMC and DA\_RMC methods on the BraTS2019 dataset under T1ce modality}
\small
\renewcommand\arraystretch{1.5}
\setlength{\tabcolsep}{1mm}{
\begin{tabular}{ccccc}
\hline
Method  & AUC   & Accuracy & Sensitivity & Specificity \\ \hline
RMC   & 0.901$\pm$0.023  & 0.911$\pm$0.031 & 0.882$\pm$0.037 & 0.919$\pm$0.041     \\
DA\_RMC & \textbf{0.930$\pm$0.028}  & \textbf{0.928$\pm$0.009} & \textbf{0.933$\pm$0.061} & \textbf{0.927$\pm$0.006}         \\ \hline
\end{tabular}
}
\label{DA_2019}
\end{table}


\subsubsection{Ablation study of guidance-aided module}
We hypothesize that the more modalities available for training, the better performance can be achieved by our multi-modal approach. For verification, we design a set of experiments under different conditions where the number and type of modalities used for building models are changed. Note that these experiments are carried out under the assumption that the primary modality(T1ce) always exists, and the availability of the secondary modalities (T1, T2, Flair) varies (i.e., the 3 secondary modalities are chosen in different combinations respectively). In total, there are $2^3=8$ types of input in total, corresponding to 8 models. The detailed results are presented in Table~\ref{tab:ablation}. The "$\checkmark$" in the table represents this modality is available in this combination, while "$-$" represents this modality is missing.


The experiment results show that with more different modalities added into the model input, the model becomes more robust, with a better performance in terms of AUC. The classification performance reaches its peak when all modalities are available. 
Sometimes, similar scores appear between different combinations, which infers that these combinations have similar complementary information beyond the primary modality. These experimental results confirm our previous assumption. 


\begin{table*}[h]
\centering
\caption{Ablation study for guidance-aided module on different situations with modality available($\checkmark$) or missing ($-$)}
\small
\renewcommand\arraystretch{1.5}
\setlength{\tabcolsep}{1.2mm}{
\begin{tabular}{cccccccccccc}
\hline
\multicolumn{4}{c}{Modalities} & \multicolumn{4}{c}{BraTS2018}                & \multicolumn{4}{c}{BraTS2019}                \\
             \cmidrule(r){1-4}       \cmidrule(r){5-8}  \cmidrule(r){9-12}
    T1ce   & Flair & T1 & T2 & AUC   & Accuracy & Sensitivity & Specificity & AUC   & Accuracy & Sensitivity & Specificity \\  \hline
$\checkmark$  & $-$  &  $-$   & $-$           & 0.941$\pm$0.020  & 0.944$\pm$0.012 & 0.932$\pm$0.049  & 0.942$\pm$0.014   & 0.930$\pm$0.028  & 0.928$\pm$0.009 & 0.933$\pm$0.061 & 0.927$\pm$0.006     \\

$\checkmark$  & $\checkmark$  &  $-$   & $-$           & 0.946$\pm$0.022  & 0.950$\pm$0.022 & 0.939$\pm$0.071  & 0.957$\pm$0.042   & 0.935$\pm$0.059  & 0.955$\pm$0.024 & 0.887$\pm$0.114  & 0.969$\pm$0.006   \\

$\checkmark$  &  $-$  & $\checkmark$  & $-$           & 0.954$\pm$0.021  & 0.954$\pm$0.025 & 0.981$\pm$0.022  & 0.957$\pm$0.042  & 0.931$\pm$0.056  & 0.949$\pm$0.019 & 0.899$\pm$0.116  & 0.962$\pm$0.005     \\ 

$\checkmark$  &  $-$   & $-$     & $\checkmark$        & 0.952$\pm$0.025  & 0.947$\pm$0.028 & 0.981$\pm$0.022  & 0.947$\pm$0.049  & 0.930$\pm$0.059  & 0.953$\pm$0.027 & 0.899$\pm$0.116  & 0.973$\pm$0.007   \\ 

$\checkmark$   & $-$   & $\checkmark$   & $\checkmark$        & 0.957$\pm$0.012  & 0.950$\pm$0.022 & 0.955$\pm$0.007  & 0.947$\pm$0.049   & 0.948$\pm$0.038  & 0.958$\pm$0.021 & 0.930$\pm$0.068 & 0.966$\pm$0.013   \\ 

$\checkmark$  & $\checkmark$  & $-$   & $\checkmark$          & 0.952$\pm$0.018  & 0.939$\pm$0.016 & 0.976$\pm$0.041  & 0.927$\pm$0.025  & 0.947$\pm$0.038  & 0.961$\pm$0.019 & 0.924$\pm$0.069 & 0.970$\pm$0.007      \\ 

$\checkmark$  & $\checkmark$   & $\checkmark$  & $-$          & 0.964$\pm$0.011  & 0.954$\pm$0.027 & 0.985$\pm$0.026  & 0.947$\pm$0.049  & 0.950$\pm$0.037  & 0.961$\pm$0.019 & 0.930$\pm$0.068  & 
{0.970$\pm$0.007}     \\ 

$\checkmark$ & $\checkmark$   & $\checkmark$   & $\checkmark$            & \textbf{0.985$\pm$0.019}      & \textbf{0.979$\pm$0.028}    & \textbf{1.000$\pm$0.000}       & \textbf{0.971$\pm$0.039}      & \textbf{0.966$\pm$0.021}  & \textbf{0.970$\pm$0.010}  & \textbf{0.958$\pm$0.039}   & \textbf{0.974$\pm$0.065}    \\

\hline
\end{tabular}
}
\label{tab:ablation}
\end{table*}

\subsection{Comparison with state-of-the-art methods}
We compared our method with three basic multi-modal fusion methods and two existing MRI multi-modal classification approaches that use different data pre-processing techniques.

The three basic multi-modal fusion methods for brain tumor grading use the pre-trained RMC model as the backbone model: 1) \textbf{pixel fusion} in which the four modalities are stacked in the channel dimension as a single input of the model; 2) \textbf{feature fusion} in which the four modalities are respectively fed into the feature extractor and their corresponding features are concatenated and then classified by the classifier; 3) \textbf{decision fusion}~\cite{amin2019new} in which all modalities are separately classified by the classifier, and their corresponding probabilities are concatenated. Then the fused results are passed into a 3-layer multi-layer perceptron to yield the final prediction. It is worth mentioning that to ensure fairness, we keep the network architecture (i.e., pre-trained RMC), and training configurations (including data pre-processing, the learning rate schedule, and the optimizer) in the three multi-modal fusion methods the same as the proposed method.

We also include two current state-of-the-art (SOTA) multi-modal approaches: i) Automated Glioma Grading (\textbf{AGG})~\cite{zhuge2020automated}, which trains a UNet-like network to perform tumor segmentation, then crops tumor regions from the MRI images based on segmentation results, and finally build a 3D CNN network to classify the ROIs; ii) Multi-modal Disentangled Variational Autoencoder (\textbf{MMD-VAE})~\cite{cheng2021multimodal}, which extracts the radiomics features from MRI images with the provided segmentation ground truth, VAEs are disentangled to obtain the similar and complementary information among modalities to make the prediction. It is worth noting that both \textbf{AGG} and \textbf{MMD-VAE} rely on tumor region annotations.

Table\ref{total} summarizes the performance of the proposed method and the compared approaches along with the baseline method (i.e., RMC model trained with T1ce) on the BraTS2018 and BraTS2019 datasets. Unfortunately, the three multi-modal fusion methods fail to achieve satisfactory results, with all AUCs below 0.9. Compared to the baseline model, there is a considerable performance drop caused by pixel fusion, feature fusion, and decision fusion. The feature and decision fusion even perform worse than the ensembling models in Table~\ref{single18} and Table~\ref{single19}, with AUCs just a little over 0.6, which infers that general multi-modal fusion might introduce more noise than complementary information. 

Surprisingly, a severe performance degradation occurs on the multi-modal method AGG model as well, which could only achieve a similar AUC to the pixel fusion algorithm on the BraTS2019 dataset (i.e 0.834 vs 0.837), but a far worse AUC on the BraTS2018 dataset (i.e., 0.758 vs 0.886), suggesting its low generalization capability. Although the MMD-VAE model has better results on the BraTS2019 dataset compared with the baseline with improvement in the AUC by 4.2\% in absolute terms, it still fails to improve the performance on the BraTS2018 dataset. More importantly, our proposed guidance-aided multi-modal framework achieves the highest scores of all evaluation metrics on the two datasets, significantly outperforming the compared methods, which proves the robustness and efficiency of our framework for brain tumor grading. It is worth mentioning that, unlike AGG and MMD-VAE which require a sophisticated data-preprocess step to acquire tumor annotations (either automated or manual segmentation) for building models, the proposed method can directly leverage the whole MRI images for training and prediction.

\begin{table*}[h]
\centering
\caption{Comparison result of our model and other grading methods on both BraTS2018 and BraTS2019. }
\small
\renewcommand\arraystretch{1.5}
\setlength{\tabcolsep}{1.2mm}{
\begin{tabular}{ccccccccc}
\hline
\multirow{2}{*}{Method} & \multicolumn{4}{c}{BraTS2018}                & \multicolumn{4}{c}{BraTS2019}                \\
                    \cmidrule(r){2-5}  \cmidrule(r){6-9}
                        & AUC   & Accuracy & Sensitivity & Specificity & AUC   & Accuracy & Sensitivity & Specificity \\  \hline
RMC (T1ce)                     & 0.917$\pm$0.061  & 0.926$\pm$0.038 & 0.933$\pm$0.046  & 0.900$\pm$0.075 & 0.901$\pm$0.023  & 0.911$\pm$0.031 & 0.882$\pm$0.037 & 0.919$\pm$0.041     \\ \hline
Pixel fusion\cite{hermessi2021multimodal}    & 0.886$\pm$0.011  & 0.868$\pm$0.021    & 0.920$\pm$0.069        & 0.853$\pm$0.047         & 0.837$\pm$0.000 & 0.901$\pm$0.000    & 0.720$\pm$0.000   & 0.954$\pm$0.000     \\
Feature fusion\cite{hermessi2021multimodal}             & 0.620$\pm$0.011  & 0.706$\pm$0.027    & 0.428$\pm$0.032        & 0.812$\pm$0.035         & 0.636$\pm$0.013 & 0.747$\pm$0.026    & 0.439$\pm$0.054   & 0.834$\pm$0.048     \\
Decision fusion\cite{amin2019new}         & 0.610$\pm$0.029  & 0.677$\pm$0.016    & 0.474$\pm$0.075        & 0.746$\pm$0.028         & 0.632$\pm$0.046 & 0.789$\pm$0.024    & 0.362$\pm$0.080   & 0.902$\pm$0.035     \\
\hline
AGG\cite{zhuge2020automated}     & 0.758$\pm$0.033  & 0.751$\pm$0.052    & 0.773$\pm$0.023        & 0.743$\pm$0.074         & 0.835$\pm$0.007 & 0.796$\pm$0.010    & 0.907$\pm$0.023   & 0.764$\pm$0.018     \\
MMD-VAE\cite{cheng2021multimodal}  & 0.905$\pm$0.082  &0.930$\pm$0.061    & 0.853$\pm$0.129        & 0.957$\pm$0.038         & {0.943$\pm$0.023} & 0.942$\pm$0.028    & {0.907$\pm$0.046}   & {0.977$\pm$0.012}     \\
Ours                    & \textbf{0.985$\pm$0.019}      & \textbf{0.979$\pm$0.028}    & \textbf{1.000$\pm$0.000}       & \textbf{0.971$\pm$0.039}      & \textbf{0.966$\pm$0.021}  & \textbf{0.970$\pm$0.010}  & \textbf{0.958$\pm$0.039}   & \textbf{0.974$\pm$0.065}     \\ \hline
\end{tabular}
}
\label{total}
\end{table*}

\subsection{Qualitative evaluation}
\label{ssec:subhead}
\begin{figure}[h]
    \centering
    \includegraphics[scale=0.47]{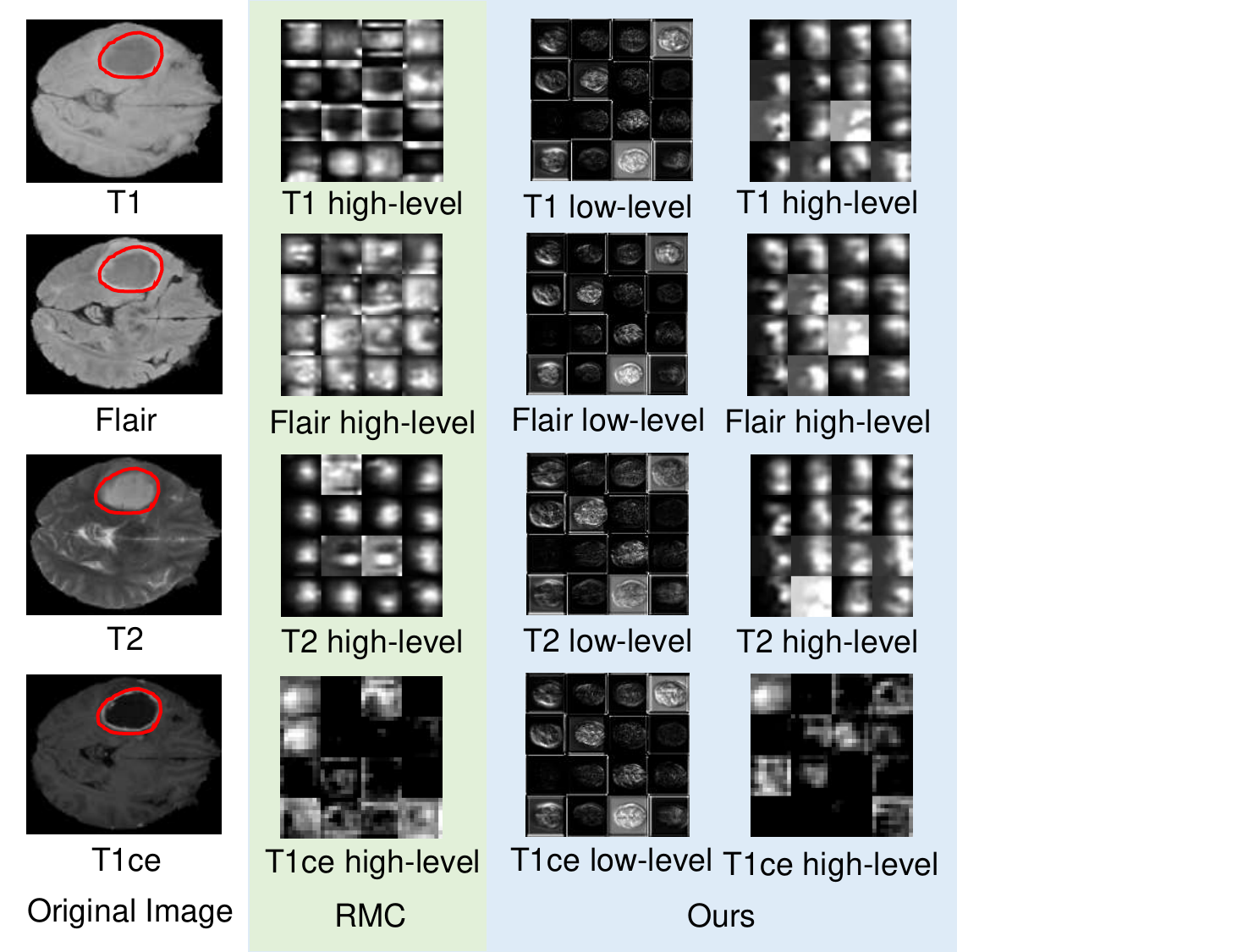}
    \caption{Feature maps from an MRI case in BraTS2019 that the RMC model predicts wrongly while our model predicts correctly. We present the original images together with different feature maps created by the SFE, LFE and HFE, these exists an obvious semantic gap between low-level features and high-level features.
    }
    \label{fig:featuremap}
\end{figure}

To verify that our guidance-aided module is truly beneficial to extract more informative features, we visualize the semantics feature maps of different modalities at different levels, which are obtained by the baseline models RMC and the proposed method. As shown in Fig~\ref{fig:featuremap}, a HGG case in BraTS2019 is selected for visualization analysis. Note that all modalities are misclassified by the baseline models. The four columns list out the original MRI images, the high-feature maps extracted by the RMC model, the low-level feature maps from our model, and the high-level feature maps from our model, respectively. A huge semantic gap between low-level and high-level features an be observed (see the last two columns). The low-level feature
contains more specific information as they are closer to the inputs, while the high-level features are more abstract, which are particularly relevant to the classification task. Compared with the high-level features generated by the RMC model (see the $2^{nd}$ column) that are blur and ambiguous, our method can capture more informative task-related features accurately from the four modalities. With the proposed cumulative learning, our method can successfully identify the HGG case. Our qualitative analysis can largely increase the interpretation ability of this work.

%% file: sections/discussion.tex
\section{Discussion}\label{discussion}
Multi-modal analysis is a promising direction for MRI brain tumor grading. By far, although some multi-modal approaches have been proposed to address this problem and shown satisfactory performance on some datasets, their generalization capability and robustness are still insufficient for real-world applications. A major drawback is that most of them highly reply on tumor region annotations, which are often difficult to obtain in practice. Besides, the approaches built based on handcrafted features suffer from limited representation ability of the features, resulting in inferior performance compared with some deep learning methods. In addition to model efficacy, the efficiency is also another important factor for model deployment in clinic. Hence, it is particularly appealing to develop a robust and efficient multi-modal framework for automated MRI brain tumor classification.

However, there are still several challenges in multi-modal analysis that should be well addressed. Multi-modal learning is not always superior to uni-modal learning, could even fail in some situations, which has been verified in our study. Specifically, we first experiment with uni-modal models to reveal the how the diagnostic information encoded in MRI images vary across different modalities, which can provide us a better understanding the characteristics of each modality. Beside, we implement some basic fusion strategies, either nonparametric (e.g., model ensemble) or parametric (e.g., the pixel, feature and decision fusion method) to leverage the multiple modalities of the MRI images, and discover that these approaches are not able to achieve better performance than the uni-modal counterparts. A possible reason is that simple modality fusion without extra regularization could introduce substantial unwanted noise that suppresses the originally informative regions, leading to information loss and performance degradation. To effectively solve this issue, we propose our guidance-aided multi-modal learning model. 
We categorize the available modalities into two groups: the primary modality and the secondary modality according to their performance by the uni-modal models. We proposed to leverage the primary modality (with best performance) provide guidance during the process of learning from the secondary modalities, with the purpose of alleviate the impact of unwanted noise and encourage model to selectively learn from the informative regions. Another merit of our method is that we build the guidance from the high-level features of the primary modality to the low-level features of each secondary modality, respectively, which could bridge the semantic gaps between the high-level features and the raw MRI images and facilitate guidance-aided feature extraction. Dual attention is also applied to enforce the model to focus on the most important features. Besides, our accumulative learning strategy could guarantee the retention of the valuable information obtained previously. Extensive experiments conducted on two public datasets (the BraTS2018 and BraTS2019 datasets) demonstrate the effectiveness of the proposed method, outperforming other multi-modal approaches significantly. We also carry out a series of ablation studies to verify the efficacy of the important components in our framework.

Despite excellent performance achieved by our method, there are still some limitations in this work. One limitation is that our framework requires two-stage training process, which might be complicated for re-implementation. Even though we adopt a lightweight backbone network in our framework to reduce the computational cost, the model size is still gigantic. Due to the bunch of parameters in the four semantics feature extractors, the training and inference process would be quite inefficient. In addition, the batch size can not be large enough, which means the strength of the batch normalization layer cannot be fully utilized. It would be interesting for future work to explore how to obtain better performance and lightweight the network structure simultaneously. 
In this work, our method is developed and tested on the the BraTS2018 and BraTS2019 datasets with and 285 and 335 samples, which might be far sufficient for a robust model. Thus, another extension of our work is to further apply our model to the large-scale datasets collected from multiple healthcare institutions. Unlike the MRI images used in this paper that are skull-stripped and have been cleaned and labeled by experts properly, the real-world data might contain more environmental noise and demonstrate a large domain shift. It is desired to leverage the advanced domain adaption techniques in future work to eliminate the domain shift issue, and make the model more robust against the noise.


%% file: sections/conclusion.tex
\section{conclusions}
\label{sec:typestyle}
In this paper, we propose an efficient guidance-aided multi-modal model for fully utilizing the complementary information from the primary modality and secondary modalities. And the dual attention module is implemented to selectively emphasize features with more diagnostic information while suppressing features with less information. The combination of these two tactics achieves satisfactory cross-modal brain tumor grading results on the BraTS2018 and BraTS2019 datasets. The three-fold cross-validation experiments result also illustrates that this algorithm is more robust and effective than the current state-of-the-art methods. Our model could achieve a pleasing result without applying any pre-processing operations, including using tumor ROI ground truth and extracting radiomics features, to the original MRI data. In future work, this guidance-aided mechanism could be further extended to other multi-modal tasks besides brain tumor grading. Moreover, with a more powerful and lightweight designed uni-modal feature extraction framework, the model will become stronger.

%% file: main.bbl
\begin{thebibliography}{10}
\providecommand{\url}[1]{#1}
\csname url@samestyle\endcsname
\providecommand{\newblock}{\relax}
\providecommand{\bibinfo}[2]{#2}
\providecommand{\BIBentrySTDinterwordspacing}{\spaceskip=0pt\relax}
\providecommand{\BIBentryALTinterwordstretchfactor}{4}
\providecommand{\BIBentryALTinterwordspacing}{\spaceskip=\fontdimen2\font plus
\BIBentryALTinterwordstretchfactor\fontdimen3\font minus
  \fontdimen4\font\relax}
\providecommand{\BIBforeignlanguage}[2]{{%
\expandafter\ifx\csname l@#1\endcsname\relax
\typeout{** WARNING: IEEEtran.bst: No hyphenation pattern has been}%
\typeout{** loaded for the language `#1'. Using the pattern for}%
\typeout{** the default language instead.}%
\else
\language=\csname l@#1\endcsname
\fi
#2}}
\providecommand{\BIBdecl}{\relax}
\BIBdecl

\bibitem{r-5}
P.~Yang, Y.~Wang, X.~Peng, G.~You, W.~Zhang, W.~Yan, Z.~Bao, Y.~Wang, X.~Qiu,
  and T.~Jiang, ``Management and survival rates in patients with glioma in
  china (2004--2010): a retrospective study from a single-institution,''
  \emph{Journal of neuro-oncology}, vol. 113, no.~2, pp. 259--266, 2013.

\bibitem{krafft2004analysis}
C.~Krafft, S.~B. Sobottka, G.~Schackert, and R.~Salzer, ``Analysis of human
  brain tissue, brain tumors and tumor cells by infrared spectroscopic
  mapping,'' \emph{Analyst}, vol. 129, no.~10, pp. 921--925, 2004.

\bibitem{lefkovits2022hgg}
S.~Lefkovits, L.~Lefkovits, and L.~Szil{\'a}gyi, ``Hgg and lgg brain tumor
  segmentation in multi-modal mri using pretrained convolutional neural
  networks of amazon sagemaker,'' \emph{Applied Sciences}, vol.~12, no.~7, p.
  3620, 2022.

\bibitem{r-6}
M.~E. Laino, R.~Young, K.~Beal, S.~Haque, Y.~Mazaheri, G.~Corrias, A.~G.
  Bitencourt, S.~Karimi, and S.~B. Thakur, ``Magnetic resonance spectroscopic
  imaging in gliomas: clinical diagnosis and radiotherapy planning,''
  \emph{BJR| Open}, vol.~2, no.~1, p. 20190026, 2020.

\bibitem{r-4}
A.~K. Altwairgi, S.~Raja, M.~Manzoor, S.~Aldandan, E.~Alsaeed, A.~Balbaid,
  H.~Alhussain, Y.~Orz, A.~Lary, and A.~A. Alsharm, ``Management and treatment
  recommendations for world health organization grade iii and iv gliomas,''
  \emph{International Journal of Health Sciences}, vol.~11, no.~3, p.~54, 2017.

\bibitem{r-2}
N.~A. Oberheim~Bush and S.~Chang, ``Treatment strategies for low-grade glioma
  in adults,'' \emph{Journal of oncology practice}, vol.~12, no.~12, pp.
  1235--1241, 2016.

\bibitem{r-3}
W.~Aiman and A.~Rayi, ``Low grade gliomas,'' \emph{StatPearls [Internet]},
  2022.

\bibitem{r-1}
D.~A. Forst, B.~V. Nahed, J.~S. Loeffler, and T.~T. Batchelor, ``Low-grade
  gliomas,'' \emph{The oncologist}, vol.~19, no.~4, pp. 403--413, 2014.

\bibitem{r-8}
T.~Reithmeier, W.~Lopez, S.~Doostkam, M.~Machein, M.~Pinsker, M.~Trippel, and
  G.~Nikkhah, ``Intraindividual comparison of histopathological diagnosis
  obtained by stereotactic serial biopsy to open surgical resection specimen in
  patients with intracranial tumours,'' \emph{Clinical Neurology and
  Neurosurgery}, vol. 115, no.~10, pp. 1955--1960, 2013.

\bibitem{r-7}
Y.~Mizobuchi, K.~Nakajima, T.~Fujihara, K.~Matsuzaki, H.~Mure, S.~Nagahiro, and
  Y.~Takagi, ``The risk of hemorrhage in stereotactic biopsy for brain
  tumors,'' \emph{The Journal of Medical Investigation}, vol.~66, no. 3.4, pp.
  314--318, 2019.

\bibitem{r-10}
S.~Bauer, R.~Wiest, L.-P. Nolte, and M.~Reyes, ``A survey of mri-based medical
  image analysis for brain tumor studies,'' \emph{Physics in Medicine \&
  Biology}, vol.~58, no.~13, p. R97, 2013.

\bibitem{r-11}
J.~Juan-Albarrac{\'\i}n, E.~Fuster-Garcia, J.~V. Manjon, M.~Robles, F.~Aparici,
  L.~Mart{\'\i}-Bonmat{\'\i}, and J.~M. Garcia-Gomez, ``Automated glioblastoma
  segmentation based on a multiparametric structured unsupervised
  classification,'' \emph{PloS one}, vol.~10, no.~5, p. e0125143, 2015.

\bibitem{cho2017classification}
H.-h. Cho and H.~Park, ``Classification of low-grade and high-grade glioma
  using multi-modal image radiomics features,'' in \emph{2017 39th Annual
  International Conference of the IEEE Engineering in Medicine and Biology
  Society (EMBC)}.\hskip 1em plus 0.5em minus 0.4em\relax IEEE, 2017, pp.
  3081--3084.

\bibitem{cheng2020prediction}
J.~Cheng, J.~Liu, H.~Yue, H.~Bai, Y.~Pan, and J.~Wang, ``Prediction of glioma
  grade using intratumoral and peritumoral radiomic features from
  multiparametric mri images,'' \emph{IEEE/ACM Transactions on Computational
  Biology and Bioinformatics}, 2020.

\bibitem{liang2018multimodal}
S.~Liang, R.~Zhang, D.~Liang, T.~Song, T.~Ai, C.~Xia, L.~Xia, and Y.~Wang,
  ``Multimodal 3d densenet for idh genotype prediction in gliomas,''
  \emph{Genes}, vol.~9, no.~8, p. 382, 2018.

\bibitem{ouerghi2022glioma}
H.~Ouerghi, O.~Mourali, and E.~Zagrouba, ``Glioma classification via mr images
  radiomics analysis,'' \emph{The Visual Computer}, vol.~38, no.~4, pp.
  1427--1441, 2022.

\bibitem{zhuge2020automated}
Y.~Zhuge, H.~Ning, P.~Mathen, J.~Y. Cheng, A.~V. Krauze, K.~Camphausen, and
  R.~W. Miller, ``Automated glioma grading on conventional mri images using
  deep convolutional neural networks,'' \emph{Medical physics}, vol.~47, no.~7,
  pp. 3044--3053, 2020.

\bibitem{shahzadi2018cnn}
I.~Shahzadi, T.~B. Tang, F.~Meriadeau, and A.~Quyyum, ``Cnn-lstm: cascaded
  framework for brain tumour classification,'' in \emph{2018 IEEE-EMBS
  Conference on Biomedical Engineering and Sciences (IECBES)}.\hskip 1em plus
  0.5em minus 0.4em\relax IEEE, 2018, pp. 633--637.

\bibitem{mai2022multimodal}
S.~Mai, Y.~Zeng, and H.~Hu, ``Multimodal information bottleneck: Learning
  minimal sufficient unimodal and multimodal representations,'' \emph{IEEE
  Transactions on Multimedia}, 2022.

\bibitem{maneesha2019multi}
P.~Maneesha, T.~Singh, R.~Nayar, and S.~Kumar, ``Multi modal medical image
  fusion using convolution neural network,'' in \emph{2019 Third International
  Conference on Inventive Systems and Control (ICISC)}.\hskip 1em plus 0.5em
  minus 0.4em\relax IEEE, 2019, pp. 351--357.

\bibitem{pei2020brain}
L.~Pei, L.~Vidyaratne, W.-W. Hsu, M.~M. Rahman, and K.~M. Iftekharuddin,
  ``Brain tumor classification using 3d convolutional neural network,'' in
  \emph{International MICCAI brainlesion workshop}.\hskip 1em plus 0.5em minus
  0.4em\relax Springer, 2020, pp. 335--342.

\bibitem{cheng2021multimodal}
J.~Cheng, M.~Gao, J.~Liu, H.~Yue, H.~Kuang, J.~Liu, and J.~Wang, ``Multimodal
  disentangled variational autoencoder with game theoretic interpretability for
  glioma grading,'' \emph{IEEE Journal of Biomedical and Health Informatics},
  vol.~26, no.~2, pp. 673--684, 2021.

\bibitem{ning2021multi}
Z.~Ning, J.~Luo, Q.~Xiao, L.~Cai, Y.~Chen, X.~Yu, J.~Wang, and Y.~Zhang,
  ``Multi-modal magnetic resonance imaging-based grading analysis for gliomas
  by integrating radiomics and deep features,'' \emph{Annals of Translational
  Medicine}, vol.~9, no.~4, 2021.

\bibitem{re-2}
J.~Seetha and S.~S. Raja, ``Brain tumor classification using convolutional
  neural networks,'' \emph{Biomedical \& Pharmacology Journal}, vol.~11, no.~3,
  p. 1457, 2018.

\bibitem{re-3}
S.~Deepak and P.~Ameer, ``Brain tumor classification using deep cnn features
  via transfer learning,'' \emph{Computers in biology and medicine}, vol. 111,
  p. 103345, 2019.

\bibitem{re-4}
F.~J. D{\'\i}az-Pernas, M.~Mart{\'\i}nez-Zarzuela, M.~Ant{\'o}n-Rodr{\'\i}guez,
  and D.~Gonz{\'a}lez-Ortega, ``A deep learning approach for brain tumor
  classification and segmentation using a multiscale convolutional neural
  network,'' in \emph{Healthcare}, vol.~9, no.~2.\hskip 1em plus 0.5em minus
  0.4em\relax MDPI, 2021, p. 153.

\bibitem{re-5}
K.~Kaplan, Y.~Kaya, M.~Kuncan, and H.~M. Ertun{\c{c}}, ``Brain tumor
  classification using modified local binary patterns (lbp) feature extraction
  methods,'' \emph{Medical hypotheses}, vol. 139, p. 109696, 2020.

\bibitem{dutta2022cdanet}
T.~K. Dutta and D.~R. Nayak, ``Cdanet: Channel split dual attention based cnn
  for brain tumor classification in mr images,'' in \emph{2022 IEEE
  International Conference on Image Processing (ICIP)}.\hskip 1em plus 0.5em
  minus 0.4em\relax IEEE, 2022, pp. 4208--4212.

\bibitem{amin2020brain}
J.~Amin, M.~Sharif, N.~Gul, M.~Yasmin, and S.~A. Shad, ``Brain tumor
  classification based on dwt fusion of mri sequences using convolutional
  neural network,'' \emph{Pattern Recognition Letters}, vol. 129, pp. 115--122,
  2020.

\bibitem{kaur2021multi}
M.~Kaur and D.~Singh, ``Multi-modality medical image fusion technique using
  multi-objective differential evolution based deep neural networks,''
  \emph{Journal of Ambient Intelligence and Humanized Computing}, vol.~12,
  no.~2, pp. 2483--2493, 2021.

\bibitem{tan2020multimodal}
W.~Tan, P.~Tiwari, H.~M. Pandey, C.~Moreira, and A.~K. Jaiswal, ``Multimodal
  medical image fusion algorithm in the era of big data,'' \emph{Neural
  Computing and Applications}, pp. 1--21, 2020.

\bibitem{wang2022disentangled}
A.~Wang, X.~Luo, Z.~Zhang, and X.-J. Wu, ``A disentangled representation based
  brain image fusion via group lasso penalty,'' \emph{Frontiers in
  Neuroscience}, vol.~16, 2022.

\bibitem{almasri2022artificial}
M.~M. Almasri and A.~M. Alajlan, ``Artificial intelligence-based multimodal
  medical image fusion using hybrid s2 optimal cnn,'' \emph{Electronics},
  vol.~11, no.~14, p. 2124, 2022.

\bibitem{hermessi2021multimodal}
H.~Hermessi, O.~Mourali, and E.~Zagrouba, ``Multimodal medical image fusion
  review: Theoretical background and recent advances,'' \emph{Signal
  Processing}, vol. 183, p. 108036, 2021.

\bibitem{liu2021contrastive}
Y.~Liu, Q.~Fan, S.~Zhang, H.~Dong, T.~Funkhouser, and L.~Yi, ``Contrastive
  multimodal fusion with tupleinfonce,'' in \emph{Proceedings of the IEEE/CVF
  International Conference on Computer Vision}, 2021, pp. 754--763.

\bibitem{hu2021mmgcn}
J.~Hu, Y.~Liu, J.~Zhao, and Q.~Jin, ``Mmgcn: Multimodal fusion via deep graph
  convolution network for emotion recognition in conversation,'' \emph{arXiv
  preprint arXiv:2107.06779}, 2021.

\bibitem{wu2021multimodal}
Y.~Wu, P.~Zhan, Y.~Zhang, L.~Wang, and Z.~Xu, ``Multimodal fusion with
  co-attention networks for fake news detection,'' in \emph{Findings of the
  Association for Computational Linguistics: ACL-IJCNLP 2021}, 2021, pp.
  2560--2569.

\bibitem{zhang2022mmformer}
Y.~Zhang, N.~He, J.~Yang, Y.~Li, D.~Wei, Y.~Huang, Y.~Zhang, Z.~He, and
  Y.~Zheng, ``mmformer: Multimodal medical transformer for incomplete
  multimodal learning of brain tumor segmentation,'' \emph{arXiv preprint
  arXiv:2206.02425}, 2022.

\bibitem{chen2016xgboost}
T.~Chen and C.~Guestrin, ``Xgboost: A scalable tree boosting system,'' in
  \emph{Proceedings of the 22nd acm sigkdd international conference on
  knowledge discovery and data mining}, 2016, pp. 785--794.

\bibitem{han2022multimodal}
Z.~Han, F.~Yang, J.~Huang, C.~Zhang, and J.~Yao, ``Multimodal dynamics:
  Dynamical fusion for trustworthy multimodal classification,'' in
  \emph{Proceedings of the IEEE/CVF Conference on Computer Vision and Pattern
  Recognition}, 2022, pp. 20\,707--20\,717.

\bibitem{wang2020makes}
W.~Wang, D.~Tran, and M.~Feiszli, ``What makes training multi-modal
  classification networks hard?'' in \emph{Proceedings of the IEEE/CVF
  Conference on Computer Vision and Pattern Recognition}, 2020, pp.
  12\,695--12\,705.

\bibitem{chatterjee2022classification}
S.~Chatterjee, F.~A. Nizamani, A.~N{\"u}rnberger, and O.~Speck,
  ``Classification of brain tumours in mr images using deep spatiospatial
  models,'' \emph{Scientific Reports}, vol.~12, no.~1, pp. 1--11, 2022.

\bibitem{vaswani2017attention}
A.~Vaswani, N.~Shazeer, N.~Parmar, J.~Uszkoreit, L.~Jones, A.~N. Gomez,
  {\L}.~Kaiser, and I.~Polosukhin, ``Attention is all you need,''
  \emph{Advances in neural information processing systems}, vol.~30, 2017.

\bibitem{fu2019dual}
J.~Fu, J.~Liu, H.~Tian, Y.~Li, Y.~Bao, Z.~Fang, and H.~Lu, ``Dual attention
  network for scene segmentation,'' in \emph{Proceedings of the IEEE/CVF
  conference on computer vision and pattern recognition}, 2019, pp. 3146--3154.

\bibitem{menze2014multimodal}
B.~H. Menze, A.~Jakab, S.~Bauer, J.~Kalpathy-Cramer, K.~Farahani, J.~Kirby,
  Y.~Burren, N.~Porz, J.~Slotboom, R.~Wiest \emph{et~al.}, ``The multimodal
  brain tumor image segmentation benchmark (brats),'' \emph{IEEE transactions
  on medical imaging}, vol.~34, no.~10, pp. 1993--2024, 2014.

\bibitem{bakas2017advancing}
S.~Bakas, H.~Akbari, A.~Sotiras, M.~Bilello, M.~Rozycki, J.~S. Kirby, J.~B.
  Freymann, K.~Farahani, and C.~Davatzikos, ``Advancing the cancer genome atlas
  glioma mri collections with expert segmentation labels and radiomic
  features,'' \emph{Scientific data}, vol.~4, no.~1, pp. 1--13, 2017.

\bibitem{bakas2018identifying}
S.~Bakas, M.~Reyes, A.~Jakab, S.~Bauer, M.~Rempfler, A.~Crimi, R.~T. Shinohara,
  C.~Berger, S.~M. Ha, M.~Rozycki \emph{et~al.}, ``Identifying the best machine
  learning algorithms for brain tumor segmentation, progression assessment, and
  overall survival prediction in the brats challenge,'' \emph{arXiv preprint
  arXiv:1811.02629}, 2018.

\bibitem{lowekamp2013design}
B.~C. Lowekamp, D.~T. Chen, L.~Ib{\'a}{\~n}ez, and D.~Blezek, ``The design of
  simpleitk,'' \emph{Frontiers in neuroinformatics}, vol.~7, p.~45, 2013.

\bibitem{kingma:adam}
D.~P. Kingma and J.~Ba, ``Adam: A method for stochastic optimization,'' in
  \emph{International Conference on Learning Representations (ICLR)}, 2015.

\bibitem{amin2019new}
J.~Amin, M.~Sharif, M.~Yasmin, T.~Saba, M.~A. Anjum, and S.~L. Fernandes, ``A
  new approach for brain tumor segmentation and classification based on score
  level fusion using transfer learning,'' \emph{Journal of medical systems},
  vol.~43, no.~11, pp. 1--16, 2019.

\end{thebibliography}
